\documentclass[sigconf,screen]{acmart}

\usepackage{multirow}
\usepackage{booktabs}
\usepackage{makecell}
\usepackage{array} 
\usepackage{balance}

\newcommand{\etal}{\textit{et al}.}
\newcommand{\ie}{\textit{i}.\textit{e}.}
\newcommand{\eg}{\textit{e}.\textit{g}.}

\AtBeginDocument{%
  }

\setcopyright{acmcopyright}
\copyrightyear{2023}
\acmYear{2023}
\setcopyright{acmlicensed}\acmConference[MM '23]{Proceedings of the 31st
ACM International Conference on Multimedia}{October 29-November 3,
2023}{Ottawa, ON, Canada}
\acmBooktitle{Proceedings of the 31st ACM International Conference on
Multimedia (MM '23), October 29-November 3, 2023, Ottawa, ON, Canada}
\acmPrice{15.00}
\acmDOI{10.1145/3581783.3612083}
\acmISBN{979-8-4007-0108-5/23/10}

\acmSubmissionID{1729}

\usepackage{subfigure}
\usepackage{booktabs} 
\usepackage{multirow}
\usepackage{color}
\begin{document}


\title{Frequency Perception Network for Camouflaged Object Detection}

\author{Runmin Cong}
\affiliation{
  \institution{ School of Control Science and Engineering, Key Laboratory of Machine Intelligence and System Control, Ministry of Education, 
  Shandong University
}
  \city{Jinan}
  \state{Shandong}
  \country{China}
}
\email{rmcong@sdu.edu.cn}

\author{Mengyao Sun}
\affiliation{
  \institution{Institute of Information Science, Beijing Key Laboratory of Advanced Information Science and Network Technology, \\Beijing Jiaotong University}
  \city{Beijing}
  \country{China}
}
\email{sunmengyao@stu.ouc.edu.cn}

\author{Sanyi Zhang}
\authornote{Corresponding author}
\affiliation{
  \institution{ State Key Laboratory of Information
Security (SKLOIS), \\Institute of Information Engineering, \\ 
  Chinese Academy of Sciences }
  \city{Beijing}
  \country{China}
}
\email{zhangsanyi@iie.ac.cn}

\author{Xiaofei Zhou}
\affiliation{
  \institution{School of Automation,\\ Hangzhou Dianzi University}
  \city{Hangzhou}
  \state{Zhejiang}
  \country{China}
}
\email{zxforchid@outlook.com}

\author{Wei Zhang}
\affiliation{
  \institution{School of Control Science and Engineering, Key Laboratory of Machine Intelligence and System Control, Ministry of Education, 
  Shandong University
}
 \city{Jinan}
 \state{Shandong}
  \country{China}
}
\email{davidzhang@sdu.edu.cn}

\author{Yao Zhao}
\affiliation{
  \institution{Institute of Information Science, Beijing Key Laboratory of Advanced Information Science and Network Technology, \\Beijing Jiaotong University}
  \city{Beijing}
  \country{China}
}
\email{yzhao@bjtu.edu.cn}

\renewcommand{\shortauthors}{Runmin Cong et al.}

\begin{abstract}
 Camouflaged object detection (COD) aims to accurately detect objects hidden in the surrounding environment. 
However, the existing COD methods mainly locate camouflaged objects in the RGB domain, their performance has not been fully exploited in many challenging scenarios.
Considering that the features of the camouflaged object and the background are more discriminative in the frequency domain, we propose a novel learnable and separable frequency perception mechanism driven by the semantic hierarchy in the frequency domain.
Our entire network adopts a two-stage model, including a frequency-guided coarse localization stage and a detail-preserving fine localization stage.
With the multi-level features extracted by the backbone, we design a flexible frequency perception module based on octave convolution for coarse positioning.
Then, we design the correction fusion module to step-by-step integrate the high-level features through the prior-guided correction and cross-layer feature channel association, and finally combine them with the shallow features to achieve the detailed correction of the camouflaged objects.
Compared with the currently existing models, our proposed method achieves competitive performance in three popular benchmark datasets both qualitatively and quantitatively. The code will be
released at \url{https://github.com/rmcong/FPNet_ACMMM23}.

\end{abstract}



\maketitle

\section{Introduction}

In nature, animals use camouflage to blend in with their surroundings to avoid detection by predators. The camouflaged object detection (COD) task aims to allow computers to automatically recognize these camouflaged objects that blend in with the background, which can be used in numerous downstream applications, including medical segmentation \cite{r24,crm/tip20/MCMT-GAN,crm/tce22/covid,crm/tim22/covid}, unconstrained face recognition \cite{chen2023weighted}, and recreational art \cite{r30,r31}.
However, the COD task is very challenging due to the low contrast properties between the camouflaged object and the background.
Furthermore, camouflaged objects may have multiple appearances, including shapes, sizes, and textures, which further increases the difficulty of detection.
\begin{figure}[!t]
\centering
\includegraphics[width=0.99\columnwidth]{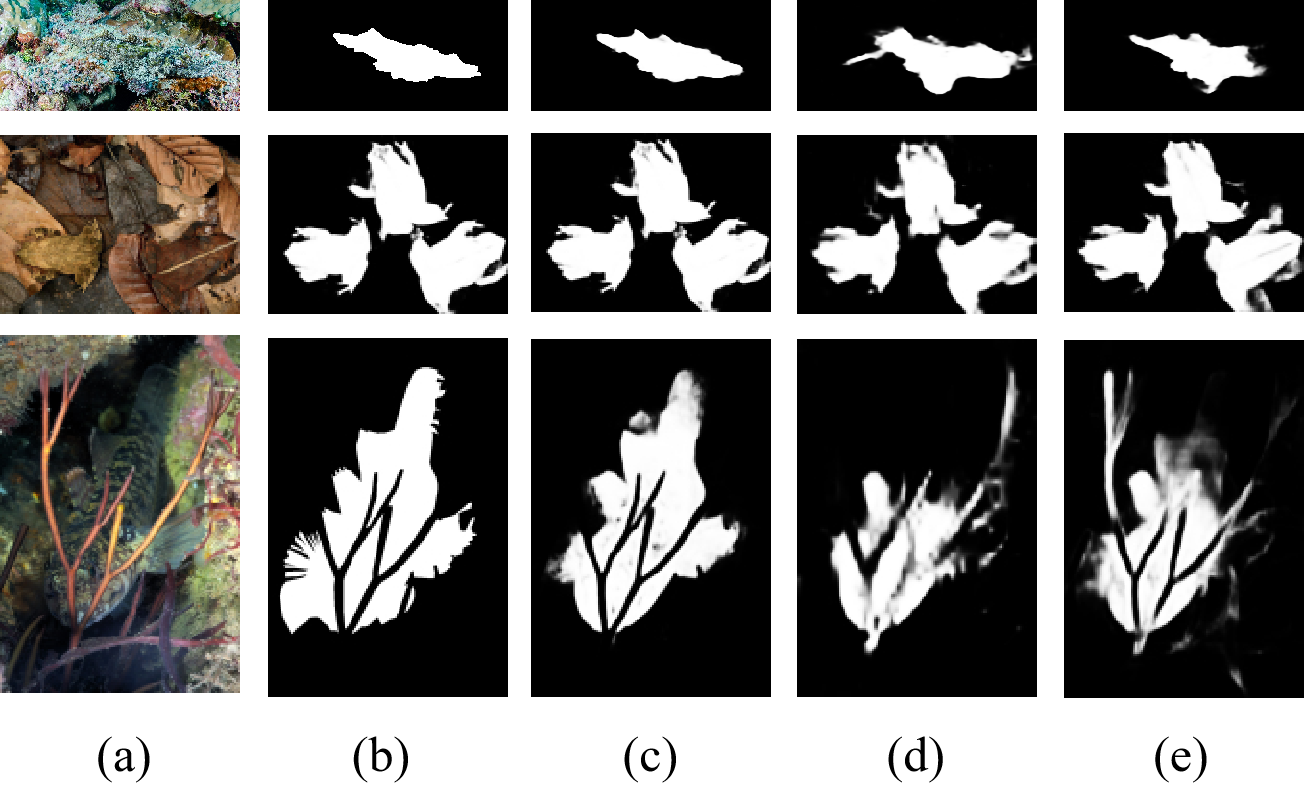}
\caption{ Three challenging camouflaged object detection (COD) scenarios from top to down are with indefinable boundaries, multiple objects, and occluded objects, respectively. The images from left to right are (a) Input image, (b) GT, (c) Ours, (d) SINet-V2 \cite{r53}, (e) LSR \cite{r38}.}
\label{fig:1}
\end{figure}

At the beginning of the research, the COD task was also regarded as a low-contrast special case of the salient object detection (SOD) task, but simple SOD model \cite{crm/tcsvt19/review,crm/aaai20/GCPANet,crm/tcsvt22/weaklySOD,crm/acmmm21/CDINet,crm/tcyb22/glnet,crm/tip22/CIRNet,crm/tnnls22/360SOD,crm/tetci22/PSNet,crm/acmmm21/light-field,crm/tmm22/TNet} retraining cannot obtain satisfactory COD results, and usually requires some special positioning design to find camouflaged objects.
Recently, with the development of deep learning \cite{r27,crm/access17/dsr,crm/spl21/underwater,crm/jbhi22/polyp, zhang2023controlvideo}, many customized networks for COD tasks have gradually emerged \cite{r37, r7, r16}. 
However, current solutions still struggle in challenging situations, such as multiple camouflaged objects, uncertain or fuzzy object boundaries, and occlusion, as shown in Figure \ref{fig:1}.
In general, these methods mainly design modules in the RGB color domain to detect camouflaged objects, and complete the initial positioning of camouflaged objects by looking for areas with inconsistent information such as textures (called breakthrough points). However, the concealment and confusion of the camouflaged objects itself make this process very difficult.
In the image frequency domain analysis, the high-frequency and low-frequency component information in the frequency domain describes the details and contour characteristics of the image in a more targeted manner, which can be used to improve the accuracy of the initial positioning.
Inspired by this, we propose a Frequency Perception Network (FPNet) that employs a two-stage strategy of search and recognition to detect camouflaged objects, taking full advantage of RGB and frequency cues.

On the one hand, the main purpose of the frequency-guided coarse positioning stage is to use the frequency domain features to find the breakthrough points of the camouflaged object position.
We first adopt the transformer backbone to extract multi-level features of the input RGB image. Subsequently, in order to realize the extraction of frequency domain features, we introduce a frequency-perception module to decompose color features into high-frequency and low-frequency components. Among them, the high-frequency features describe texture features or rapidly changing parts, while the low-frequency features can outline the overall contour of the image. Considering that both texture and contour are important for camouflaged object localization, we fuse them as a complete representation of frequency domain information. In addition, a neighbor interaction mechanism is also employed to combine different levels of frequency-aware features, thereby achieving coarse detection and localization of camouflaged objects.
On the other hand, the detail-preserving fine localization stage focuses on progressively prior-guided correction and fusion across layers, thereby generating the final finely camouflaged object masks. Specifically, we design the correction fusion module to achieve the cross-layer high-level feature interaction by integrating the prior-guided correction and cross-layer feature channel association.
Finally, the shallow high-resolution features are further introduced to refine and modify the boundaries of camouflaged objects and generate the final COD result.

The main contributions are summarized as follows:
\begin{itemize}
    \item We propose a novel two-stage framework to deeply exploit the advantages of RGB and frequency domains for camouflaged object detection in an end-to-end manner. The proposed network achieves competitive performance on three popular benchmark datasets (\ie, COD10K, CHAMELEON, and CAMO).
\end{itemize}
\begin{itemize}
    \item A novel fully frequency-perception module is designed to enhance the ability to distinguish camouflaged objects from backgrounds by automatically learning high-frequency and low-frequency features, thereby achieving coarse localization of camouflaged objects.
\end{itemize}
\begin{itemize}
    \item We design a progressive refinement mechanism to obtain the final refined camouflaged object detection results through prior-guided correction, cross-layer feature channel association, and shallow high-resolution boundary refinement.
\end{itemize}

\section{Related Work}
The COD task aims to localize objects that have a similar appearance to the background, which makes it extremely challenging. Early methods employed hand-crafted low-level features to achieve this goal, such as color \cite{r2}, expectation-maximization statistics \cite{r13}, convex intensity \cite{r5}, optical flow \cite{r14}, and texture \cite{r1,r3}. However, due to the imperceptible differences between objects and backgrounds in complex environments, and the limited expressive power of hand-crafted features, they do not perform satisfactorily.

Recently CNN-based methods \cite{r41,r7,r15} have achieved significant success in the COD task. 
In general, CNN-based methods often employ one or more of the following strategies, such as two-stage strategy \cite{r7,r16}, multi-task learning strategy \cite{r38}, and incorporating other guiding cues such as frequency \cite{r56}. 
For instance, Fan \etal~\cite{r7} proposed a two-stage process named SINet, which represents the new state-of-the-art in existing COD datasets and created the largest COD10K dataset with 10K images. 
Mei \etal~\cite{r15} imitated the predator-prey process in nature and developed a two-stage bionic framework called PFNet.

\begin{figure*}[t]
\centering
\includegraphics[width=1\linewidth]{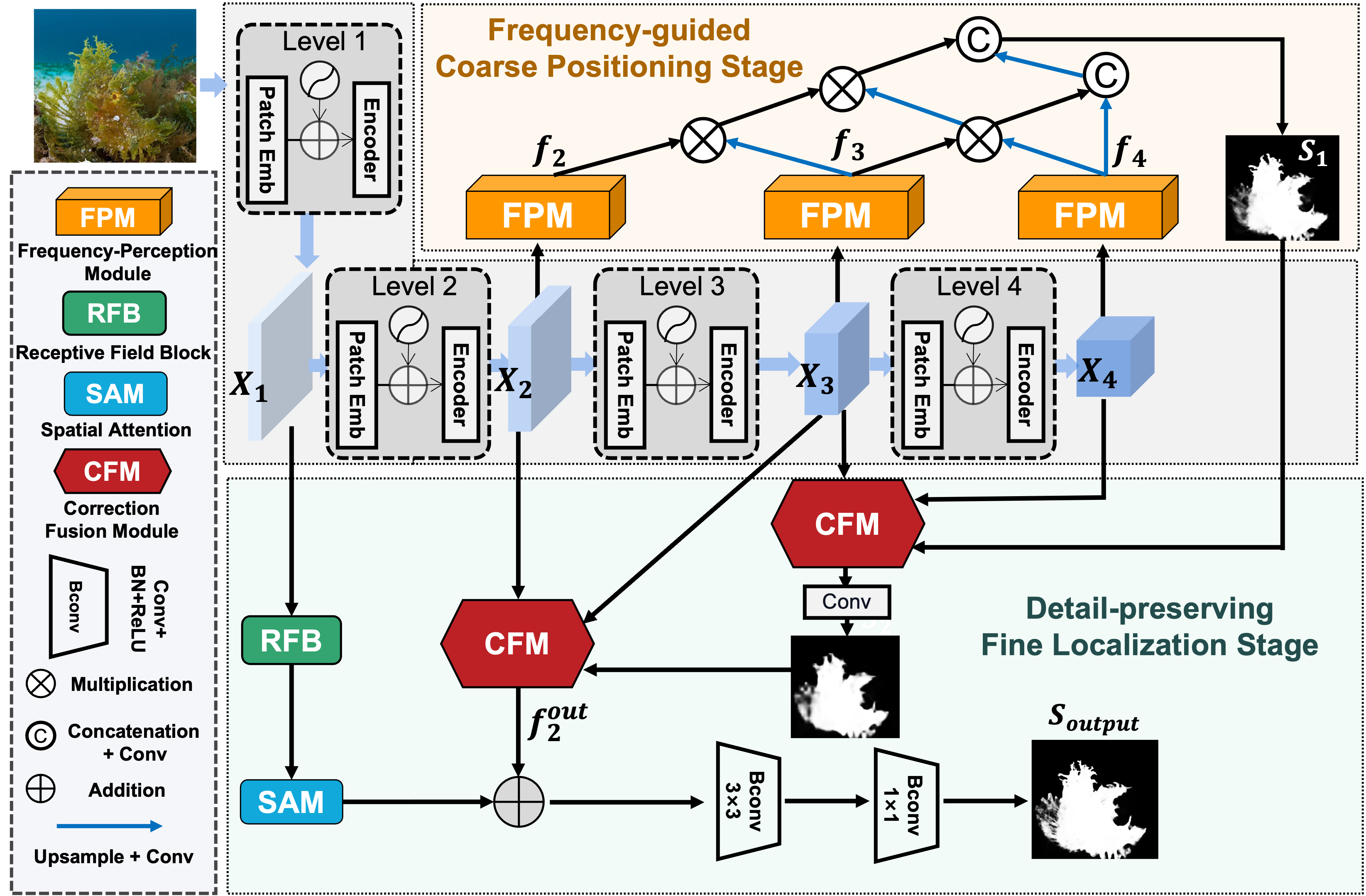}
\caption{The overview of our proposed two-stage network FPNet. The input image is first extracted with multi-level features by a PVT encoder. In the frequency-guided coarse localization stage, we use FPM for frequency-domain feature extraction and generate the coarse COD map $S_1$. Then, in the detail-preserving fine localization stage, the CFM is used to achieve progressively prior-guided correction and fusion across high-level layers. Finally, the first-level high-resolution features are further introduced to refine the boundaries of camouflaged objects and generate the final result $S_{output}$.
}
\label{fig:2}
\end{figure*}

In terms of frequency domain studies, Gueguen \etal~\cite{r56} directly used the Discrete Cosine Transform (DCT) coefficients of the image as input to CNN for subsequent visual tasks. 
Ehrlich \etal~\cite{r57} presented a general conversion algorithm for transforming spatial domain networks to the frequency domain. Interestingly, both of these works delve deep into the frequency domain transformation of the image JPEG compression process. 
Subsequently, Zhong \etal~\cite{r39} modeled the interaction between the frequency domain and the RGB domain, introducing the frequency domain as an additional cue to better detect camouflaged objects from backgrounds. 
Unlike these methods, on the one hand, we use octave convolution to realize the online learning of frequency domain features, instead of offline extraction methods (\eg, DCT); on the other hand, frequency domain features are mainly used for coarse positioning in the first stage, that is, by making full use of high-frequency and low-frequency information to find the breakthrough point of camouflaged object positioning in the frequency domain.

In addition, some methods \cite{r10,r11,r12} also try to combine edge detection to extract more precise edges, thereby improving the accuracy of COD. It is worth mentioning that in order to exploit the power of the Transformer model in the COD task, many Transformer-based methods have emerged.
For example, Yang \etal~\cite{r17} proposed to incorporate Bayesian learning into Transformer-based reasoning to achieve the COD task. The T2Net proposed by Mao \etal~\cite{r61} in 2021 used a Swin-Transformer as the backbone network, surpassing all CNN-based approaches at that time.

\section{Our Approach}
\subsection{Overview}
Our goal is to exploit and fuse the inherent advantages of the RGB and frequency domains to enhance the discrimination ability to discover camouflaged objects in the complex background. To that end, in this paper, we propose a Frequency Perception Network (FPNet) for camouflaged object detection, as shown in Figure \ref{fig:2}, including a feature extraction backbone, a frequency-guided coarse localization stage, and a detail-preserving fine localization stage.

Given an input image \(I\in \mathbb{R}^{H\times W\times 3}\), for the feature extraction backbone, we adopt the Pyramid Vision Transformer (PVT) \cite{r19} as the encoder to generate features of different levels, denoted as $X_i$ $(i= \{1,2,3,4\})$.
Each feature map serves a different purpose. The first-level feature map $X_1$ includes rich detailed information about the camouflaged object, whereas the deeper-level features (\(X_2\), \(X_3\), \(X_4\)) contain higher-level semantic information. 
With the pyramid backbone features, in the frequency-guided coarse localization stage, we first use a frequency-perception module (FPM) for frequency-domain feature extraction on high-level features and then adopt the neighborhood connection decoder for feature fusion decoding to obtain the coarse COD map $S_1$.
Whereafter, in the detail-preserving fine localization stage, with the guidance of coarse COD map $S_1$, the high-level features are embedded into the correction fusion module (CFM) to progressively achieve prior-guided correction and fusion across layers. 
Finally, a receptive field block (RFB) with spatial attention mechanism (SAM) is used for low-level high-resolution feature optimization and combined with the CFM module output to obtain the final COD result $S_{output}$.

\subsection{Frequency-guided Coarse Positioning}

\begin{figure}[!t]
\centering
\includegraphics[width=1\columnwidth]{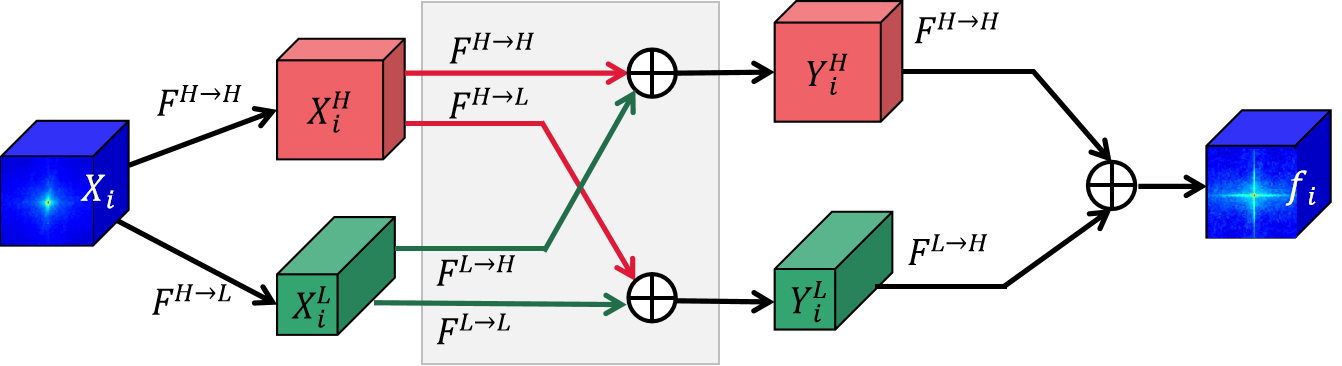}
\caption{Illustration of frequency-perception module (FPM). Two branches are for high-frequency and low-frequency information learning, respectively. }
\label{fig:4}
\end{figure}

Inspired by predator hunting systems, frequency information is more advantageous than RGB appearance features for a specific predator in the wild environment. 
This point of view has also been verified in \cite{r39}, and then a frequency domain method for camouflaged object detection is proposed.
Specifically, this work \cite{r39} used offline discrete cosine transform to convert the RGB domain information of an image to the frequency domain, but the offline frequency extraction method limits its flexibility.
As described in \cite{r40}, octave convolution can learn to divide an image into low and high frequency components in the frequency domain. 
The low-frequency features correspond to pixel points with gentle intensity transformations, such as large color blocks, that often represent the main part of the object. The high-frequency components, on the other hand, refer to pixels with intense brightness changes, such as the edges of objects in the image.
Inspired by this, we propose a frequency-perception module to automatically separate features into high-frequency and low-frequency parts, and then form a frequency-domain feature representation of camouflaged objects, the detailed process is shown in Figure \ref{fig:4}.

Specifically, we employ octave convolution \cite{r40} to automatically perceive high-frequency and low-frequency information in an end-to-end manner, enabling online learning of camouflaged object detection. The octave convolution can effectively avoid blockiness caused by the DCT and utilize the advantage of the computational speed of GPUs. In addition, 
it can be easily plugged into arbitrary networks. 
The detailed process of output of the octave convolution \(Y_i=\{Y_i^H,Y_i^L\}\) could be described in the following:  
\begin{equation}
    Y_i^H=F(X_i^H;W^{H\rightarrow H})+\mathrm{Upsample}(F(X_i^L;W^{L\rightarrow H}),2),
    \label{eq1}
\end{equation}
\begin{equation}
    Y_i^L=F(X_i^L;W^{L\rightarrow L})+F(\mathrm{pool}(X_i^H,2);W^{H\rightarrow L}),
    \label{eq2}
\end{equation}
where \(F(X;W)\) denotes a convolution with the learnable parameters of \(W\), \(\mathrm{pool}(X, k)\) is an average pooling operation with kernel size of $k \times k$, and \(\mathrm{Upsample}(X, s)\) is an up-sampling operation by a factor of \(s\) via nearest interpolation.

Considering that both high-frequency texture attribute and low-frequency contour attribute are important for camouflaged object localization, we fuse them as a complete representation of frequency domain information:
\begin{equation}
    f_i= \mathrm{Resize}(Y_i^H) \oplus \mathrm{Resize}(Y_i^L),
    \label{eq2-2}
\end{equation}
where $\mathrm{Resize}$ means to adjust features to a fixed dimension, and $\oplus$ is the element-wise addition.


Then, the Neighbor Connection Decoder (NCD) \cite{r53}, as shown in the top region (the part above the three FPMs) of Figure \ref{fig:2}, is adopted to gradually integrate the frequency-domain features of the top-three layers, fully utilizing the cross-layer semantic context relationship through the neighbor layer connection, which can be represented as:

\begin{equation}
\left\{
             \begin{array}{lr}
             f_4'=\mathscr{g^2_\uparrow}(f_4), &  \\
             f_3'=f_3 \otimes \mathscr{g^2_\uparrow}(f_4)\\
             f_2'= \mathrm{cat}(f_2 \otimes \mathscr{g^2_\uparrow} (f_3'),\mathrm{cat}(f_3',f_4')), &  
             \end{array}
\right.
    \label{eq3}
\end{equation}
where \(\otimes\) is element-wise multiplication, \(\mathscr{g^2_\uparrow}(x)\) denotes an up-sampling along with a $3 \times 3$ convolution, $\mathrm{cat}()$ denotes concatenation along with a $3 \times 3$ convolution, and \(f_2'\)  is the output of NCD. 
After this stage, we use a simple convolution to obtain a coarse mask $S_1$ that reveals the initial location of the camouflaged object.

\subsection {Detail-preserving Fine Localization}

In the previous section, we introduced how to use frequency-domain features to achieve coarse localization of camouflaged objects. But the first stage is more like a process of finding and locating breakthrough points, the integrity and accuracy of results are still not enough. To this end, we propose a detail-preserving fine localization mechanism, which not only achieves a progressive fusion of high-level features through prior correction and channel association but also considers high-resolution features to refine the boundaries of camouflaged objects, as shown in Figure \ref{fig:2}.

\begin{figure}[!t]
\centering
\includegraphics[width=1\columnwidth]{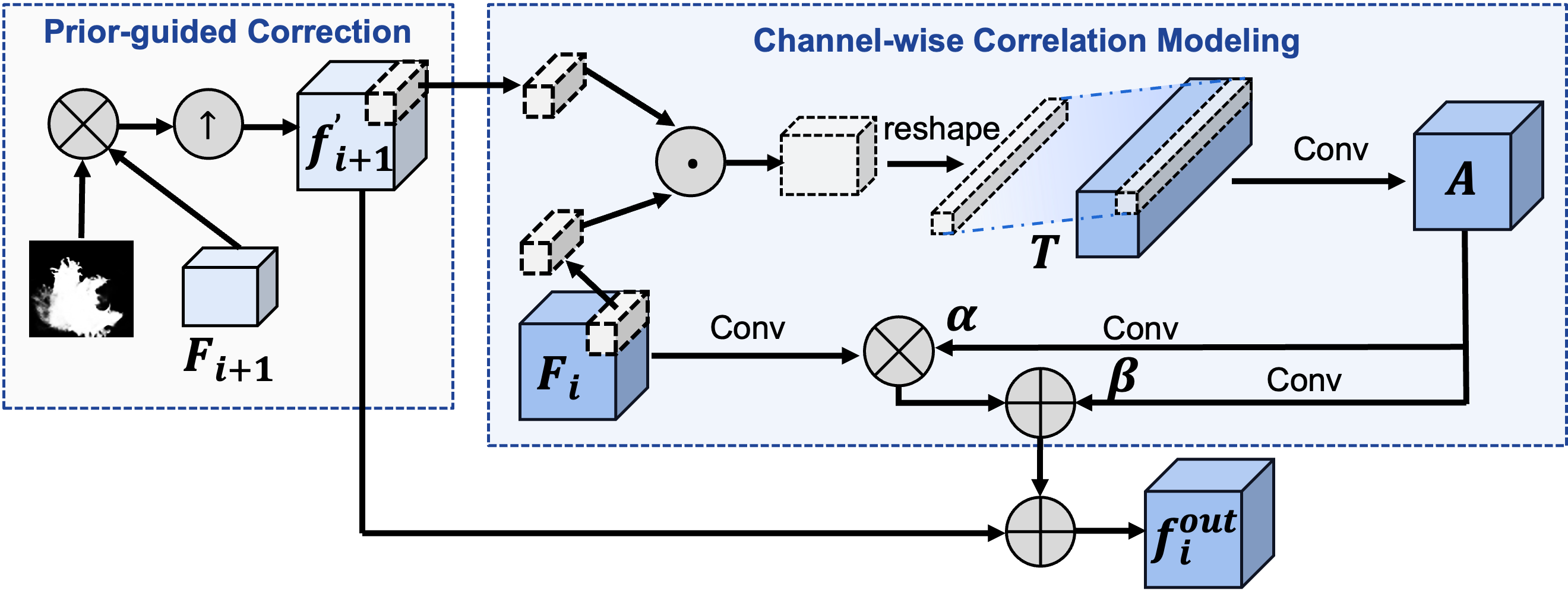}
\caption{The schematic illustration of the correction fusion module (CFM). CFM contains two parts, \ie, prior-guided correction and channel-wise correlation modeling. }
\label{fig:5}
\end{figure}


To achieve the above goals, we first design a correction fusion module (CFM), which effectively fuses adjacent layer features and a coarse camouflaged mask to produce fine output. The module includes three inputs: the current and previous layer features \(X_{i}\) and \(X_{i+1}\), and the coarse mask $S_g=\{S_1,S_2\}$. In addition, we first reduce the number of input feature channels to $64$, denoted as $F_i$ and $F_{i+1}$, which helps to improve computational efficiency while still retaining relevant information for detection.  
As shown in Figure \ref{fig:5}, our CFM consists of two parts.
In order to make full use of the existing prior guidance map \(S_g\), we purify the features of the previous layer and select the features most related to the camouflaged features to participate in the subsequent cross-layer interaction.
Mathematically, the feature map \(F_{i+1}\) is first multiplied with the coarse mask \(S_g\) to obtain the output features \(f^{'}_{i+1}\): 
\begin{equation}
    f^{'}_{i+1} = \mathrm{Upsample}(F_{i+1} \odot S_g),
    \label{eq4}
\end{equation}
where $\odot$ denotes element-wise multiplication, and $\mathrm{Upsample}$ is the upsampling operation. 
This prior-guided correction is particularly beneficial in scenarios where the object is difficult to discern from its surroundings.

It is well known that high-level features possess very rich channel-aware cues. In order to achieve more sufficient cross-layer feature interaction and effectively transfer the high-level information of the previous layer to the current layer, we design the channel-level association modeling.
We perform channel attention by taking the inner product between each pixel point on \(F_i\) and \(f^{'} _{i+1}\), which calculates the similarity between different feature maps in the channel dimension of the same pixel. 
To further reduce computational complexity, we also employ a \(3\times 3\) convolution that creates a bottleneck structure, thereby compressing the number of output channels. This process can be described as:  
\begin{equation}
A = \mathrm{conv}(F_i \otimes (f_{i+1}')^T),
    \label{eq6}
\end{equation}
where $\otimes$ is the matrix multiplication.
Then, we learn two weight maps, \(\alpha\) and \(\beta\), by using two \(3\times 3\) convolution operations on the features \(A\). 
They are further used in the correction of the features of the current layer \(F_i\) in a modulation manner. 
In this way, the final cross-level fusion features can be generated through the residual processing:    
\begin{equation}
    f^{out}_i = f_{i+1}^{'} + \mathrm{conv} (F_i) *\alpha +\beta.
\end{equation}

\begin{table*}[!t]
\centering
\caption{Comparisons of state-of-the-art methods on COD datasets. The top three results are highlighted in 
\textcolor{red}{red}, \textcolor{green}{green}, and \textcolor{blue}{blue}, respectively. 
}
\begin{tabular}{l|l|llll|llll|llll}
\hline
\multirow{2}{*}{Methods} & \multirow{2}{*}{Year} & \multicolumn{4}{c|}{COD10K-Test (2026 images)} & \multicolumn{4}{c|}{CAMO-Test (250 images)} & \multicolumn{4}{c}{CHAMELEON (76 images)} \\ \cline{3-14} 
                         & &   \(S_\alpha\uparrow\)       & \(E_{mean} \uparrow\)      &  \(F_\beta^\omega \uparrow\)       &  \(M\downarrow\)       & \(S_\alpha\uparrow\)      & \(E_{mean}\uparrow\)       & \(F_\beta^\omega \uparrow\)        &  \(M\downarrow\)    &  \(S_\alpha\uparrow\)    &  \(E_{mean} \uparrow\)       &\(F_\beta^\omega \uparrow\)        & \(M\downarrow\)      \\ \hline
FPN      {\cite{r58}}        & 2017-CVPR             & 0.697     & 0.691     & 0.411     & 0.075    & 0.684    & 0.677    & 0.483    & 0.131    & 0.794    & 0.783    & 0.590   & 0.075   \\
MaskRCNN {\cite{r59}}        & 2017-ICCV             & 0.613     & 0.748     & 0.402     & 0.080    & 0.574    & 0.715    & 0.430    & 0.151    & 0.643    & 0.778    & 0.518   & 0.099   \\
CPD      {\cite{r34}}        & 2019-CVPR             & 0.747     & 0.770     & 0.508     & 0.059    & 0.726    & 0.729    & 0.550    & 0.115    & 0.853    & 0.866    & 0.706   & 0.052   \\
SINet     {\cite{r7}}        & 2020-CVPR             & 0.771     & 0.806     & 0.551     & 0.051    & 0.751    & 0.771    & 0.606    & 0.100    & 0.869    & 0.891    & 0.740   & 0.044   \\
PraNet    {\cite{r24}}           & 2020-MICCAI           & 0.789     & 0.857     & 0.608     & 0.047    & 0.774    & 0.828    & 0.680    & 0.094    & 0.871    & 0.924    & 0.758   & 0.037   \\
PFNet     {\cite{r15}}          & 2021-CVPR             & 0.798     & 0.874     & 0.646     & 0.040    & 0.773    & 0.829    & 0.703    & 0.086    & 0.878    & 0.921    & 0.796   & 0.034   \\
C2FNet    {\cite{r52}}            & 2021-IJCAI            & 0.809     & 0.884     & 0.662     & 0.038    & 0.787    & 0.840    & 0.716    & 0.085    & 0.892    & 0.946    & 0.819   & 0.030   \\
UGTR      {\cite{r17}}            & 2021-ICCV             & 0.818     & 0.850     & 0.667     & 0.035    & 0.785    & 0.859    & 0.686    & 0.086    & 0.888    & 0.918    & 0.796   & 0.031   \\
LSR       {\cite{r38}}           & 2021-CVPR             & 0.767     & 0.861     & 0.611     & 0.045    & 0.712    & 0.791    & 0.583    & 0.104    & 0.846    & 0.913    & 0.767   & 0.046   \\
SINet-V2  {\cite{r53}}            & 2022-TPAMI             & 0.815     & 0.886     & 0.664     & 0.036    & 0.809    & 0.864    & 0.729    & 0.073    & 0.888    & 0.940    & 0.797   & 0.029   \\
FreNet    {\cite{r39}}            & 2022-CVPR             & \color{blue}0.833     &\color{blue} 0.907     & \color{blue}0.711     &\color{blue} 0.033    &\color{green} 0.828    & \color{blue}0.884    & \color{blue}0.747    &\color{blue} 0.069    & \color{blue}0.894    & \color{blue}0.950    & \color{blue}0.819   & \color{blue}0.030   \\
ZoomNet   {\cite{r55}}            & 2022-CVPR             & \color{green}0.838     &\color{green} 0.911     & \color{green}0.729     &\color{red} 0.029    & \color{blue}0.820    &\color{green} 0.892    & \color{green}0.752    & \color{green}0.066    & \color{green}0.902    & \color{green}0.958    &\color{green} 0.845   & \color{green}0.023   \\ \hline
FPNet (ours)                   & 2023                  & \color{red}0.850     &\color{red} 0.913     &\color{red} 0.748     & \color{red}0.029    & \color{red}0.852    & \color{red}0.905    & \color{red}0.806    &\color{red} 0.056    & \color{red}0.914    &\color{red} 0.961    & \color{red}0.856   &\color{red} 0.022   \\ \hline
\end{tabular}
\label{tab:all}
\end{table*}

In addition to the above-mentioned prior correction and channel-wise association modeling on the high-level features, we also make full use of the high-resolution information of the first layer to supplement the detailed information.
Specifically, we use the receptive field block (RFB) module \cite{r60} and the spatial attention module \cite{r63} on the first-layer features ($X_1$) to enlarge the receptive field and highlight the important spatial information of the features, and then fuse with the output of the CFM module ($f^{out}_2$) to generate the final prediction map:
\begin{equation}
    S_{output} = Bconv(Bconv(SAM(RFB(X_1))\oplus f^{out}_2)),
\end{equation}
where $RFB$ and $SAM$ are the receptive field block and the spatial attention module, respectively. $Bconv$ represents the $3 \times 3$ convolution layer along with the batch normalization and ReLU.

\subsection {Loss Function}
Following \cite{r51}, We compute the weighted binary cross-entropy loss \((\mathcal{L}_{BCE}^\omega)\) and IoU loss \((\mathcal{L}_{IoU}^\omega)\) on three COD maps (\ie, $S_1$, $S_2$, and $S_{output}$) to form our final loss function:
\begin{equation}
    \mathcal{L}_{total} = \mathcal{L}_1 + \mathcal{L}_2 + \mathcal{L}_{output},
\end{equation}
where $\mathcal{L}_* = \mathcal{L}_{BCE}^\omega +\mathcal{L}_{IoU}^\omega$, $*=\{1, 2, output\}$, \(\mathcal{L}_1\) denotes the loss between the coarse prediction map $S_1$ and ground truth, \(\mathcal{L}_2\) denotes the loss about the prediction map $S_2$ after the first CFM, and \(\mathcal{L}_{output}\) denotes the loss between the final prediction map $S_{output}$ and ground truth.

\section{Experiment}

\subsection {Experimental Settings}
\textbf{Datasets. }
We conduct experiments and evaluate our proposed method on three popular benchmark datasets, \textit{i.e.}, CHAMELEON \cite{r42}, CAMO \cite{r41}, and COD10K \cite{r7}. 
CHAMELEON \cite{r42} dataset has 76 images. CAMO \cite{r41} contains 1,250 camouflaged images covering different categories, which are divided into 1,000 training images and 250 testing images, respectively. As the largest benchmark dataset currently, COD10K \cite{r7} includes 5,066 images in total, 3,040 images are chosen for training and 2,026 images are used for testing. There are five concealed super-classes \textit{(\ie, terrestrial, atmobios, aquatic, amphibian, other)} and 69 sub-classes. And the pixel-level ground-truth annotations of each image in these three datasets are provided. Besides, for a fair comparison, we follow the same training strategy of previous works \cite{r39}, our training set includes 3,040 images from COD10K datasets and 1,000 images from the CAMO dataset. 

\noindent \textbf{Evaluation Metrics. }
We use four widely used and standard metrics to evaluate the proposed method, \textit{\ie,} structure-measure \((S_\alpha)\) \cite{r43}, mean E-measure \((E_\phi)\) \cite{r45}, weighted F-measure \((F_\beta ^\omega)\) \cite{r44}, and mean absolute error \((MAE)\) \cite{crm/eccv20/RGBDSOD,crm/nips20/CoADNet,crm/tcyb21/ASIFNet}. Overall, a better COD method has larger \(S_\alpha\), \(E_\phi\), and \(F_\beta ^\omega\) scores, but a smaller \(MAE\) score.

\noindent \textbf{Implementation Details. }
In this paper, we propose a frequency-perception network (FPNet) to address the challenge of camouflaged object detection by incorporating both RGB and frequency domains. Specifically, a frequency-perception module is proposed to automatically separate frequency information leading the model to a good coarse mask at the first stage. Then, a detail-preserving fine localization module equipped with a correction fusion module is explored to refine the coarse prediction map. Comprehensive comparisons and ablation studies on three benchmark COD datasets have validated the effectiveness of the proposed FPNet. 
The proposed method is implemented with PyTorch and leverages Pyramid Vision Transformer \cite{r19} pre-trained on ImageNet \cite{r49} as our backbone network. We also implement our network by using the MindSpore Lite tool\footnote{\url{https://www.mindspore.cn/}}. To update the network parameters, we use the Adam optimizer, which is widely used in transformer-based networks \cite{r19,r20,r21}. The initial learning rate is set to 1e-4 and weight decay is adjusted to 1e-4. Furthermore, we resize the input images to $512 \times 512$, the model is trained with a mini-batch size of 4 for 100 epochs on an NVIDIA 2080Ti GPU. 
We augment the training data by applying techniques such as random flipping, random cropping, and so on. 

\subsection {Comparison with State-of-the-art Methods}
\begin{figure}[!t]
\centering
\includegraphics[width=1\linewidth]{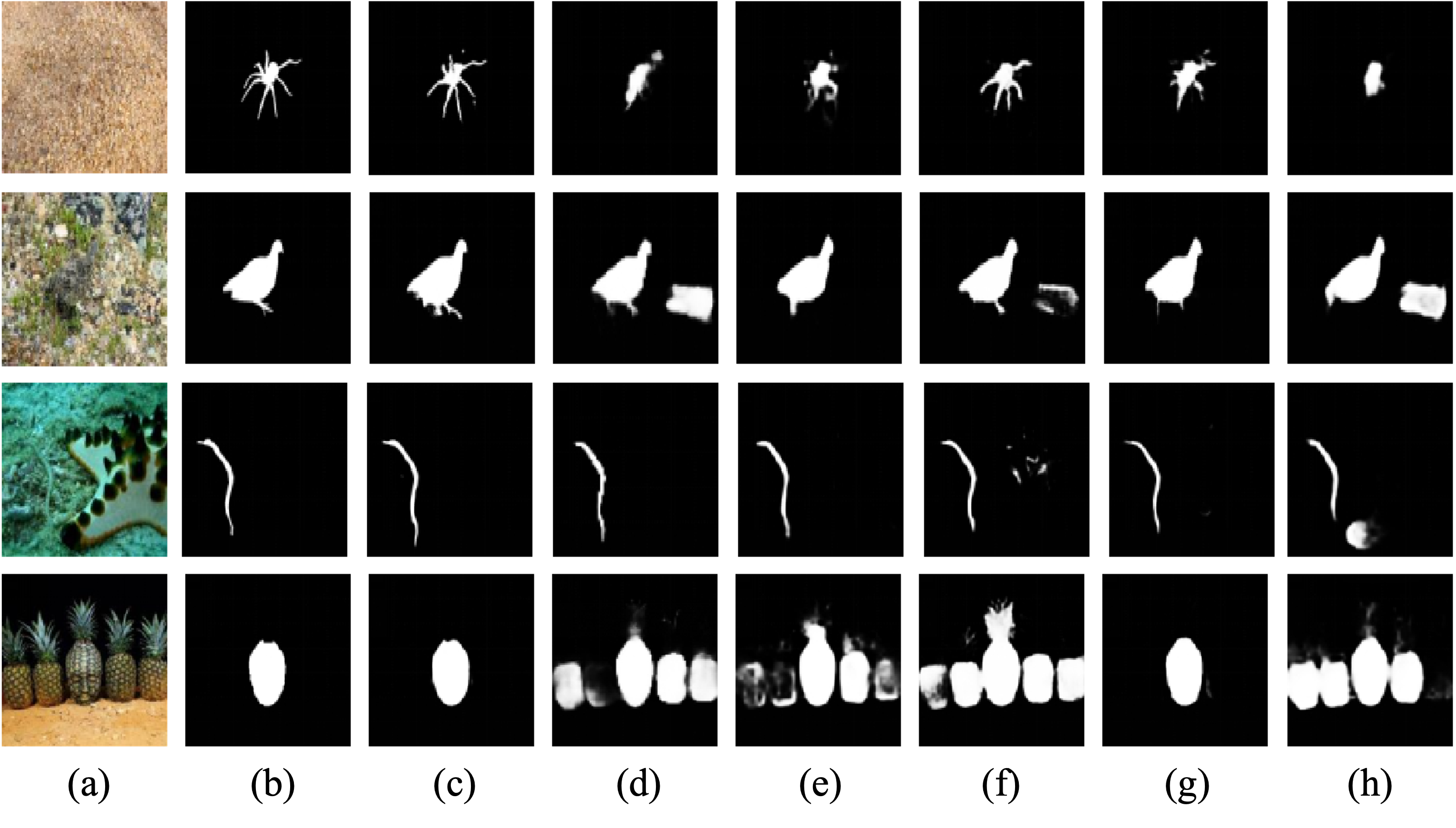}
\caption{Qualitative results of our proposed FPNet model and some state-of-the-art COD methods. The images from left to right are (a) Input image, (b) GT, (c) Ours, (d) FreNet \cite{r39}, (e) SINet-V2 \cite{r53}, (f) PFNet \cite{r15}, (g) LSR \cite{r38}, and (h) PraNet \cite{r24}.}
\label{fig:6}
\end{figure}

We conduct a comparison of our proposed method with 12 state-of-the-art mthods, including FPN      \cite{r58}, MaskRCNN \cite{r59}, CPD \cite{r34}, SINet \cite{r7}, LSR \cite{r38}, PraNet \cite{r24}, C2FNet \cite{r52}, UGTR \cite{r17}, PFNet \cite{r15}, ZoomNet \cite{r55}, SINet-V2 \cite{r53}, and FreNet \cite{r39}. 
The visualization comparisons and quantitative results are shown in Figure \ref{fig:6}, and Table \ref{tab:all} summarizes the quantitative results of the COD methods on three benchmark datasets.

\noindent \textbf{Quantitative Evaluation. }
Table \ref{tab:all} presents a detailed comparison of evaluation metrics, we can observe that our proposed model (FPNet) outperforms all SOTA models on all datasets. 
For example, our FPNet achieves obvious performance gains over other state-of-the-art ones on the CAMO-Test dataset. According to Table \ref{tab:all}, our proposed FPNet achieves the best weighted F-measure \((F_\beta ^\omega)\) score of 0.806 on the CAMO-Test dataset, and the MAE score outperforms the second-best method ZoomNet \cite{r55} by 15.2\%. 
Moreover, the proposed FPNet outperforms ZoomNet \cite{r55} by an obvious margin in terms of the \(F_\beta ^\omega\) on all three datasets. 
For example, compared with the second-best method, the percentage gain of the $F_{\beta}^{\omega}$ reach 2.6\%, 7.2\%, and 1.3\% on the COD10K-Test, CAMO-Test and CHAMELEON datasets, respectively.
While we observe the frequency-guided method FreNet \cite{r39}, it can achieve better performance than most state-of-the-art methods.
However, our proposed FPNet outperforms FreNet comprehensively in terms of all evaluation metrics, indicating that the proposed learnable frequency-guided solution is superior in discerning discriminative cues of camouflaged objects.

\noindent \textbf{Qualitative Evaluation. }
As shown in Figure \ref{fig:6}, whether the camouflaged object in the image is terrestrial or aquatic, or a camouflaged human, the proposed FPNet method is capable of accurately predicting the region of the camouflaged object. 
When the camouflaged object is extremely similar to the background, as illustrated in the first row in Figure \ref{fig:6}, other SOTA methods fail to accurately distinguish the camouflaged object, especially on the edge regions. By contrast, the proposed FPNet, benefiting from a frequency-aware learning mechanism, can clearly predict the mask of objects with clear and sharp boundaries.
When tackling complex background interference, including the salient but non-camouflaged objects (see the third row of Figure \ref{fig:6}), our proposed FPNet is capable of effectively separating the camouflaged object from the background, with a more complete and clear structure description ability.
For the approximate appearance of similar objects, as shown in the fourth row of Figure \ref{fig:6}, the camouflaged human face is hard to distinguish from other pineapples. Most methods fail to recognize it, but our proposed FPNet can discern it clearly. 
Additionally, our proposed FPNet is also effective in detecting some challenging situations (as displayed in Figure \ref{fig:1}), such as indefinable boundaries, multiple objects and occluded objects. The impressive prediction results further highlight the usefulness of the frequency-perception mechanism which connects RGB-aware and frequency-aware clues together to arrive at a unified solution that can adaptly address challenging scenarios.

\subsection {Ablation Studies}

\subsubsection{\textbf{The effectiveness of each module of FPNet}}

To verify the effectiveness of the proposed network, we separate FPNet into a series of ablation parts, \textit{i.e.},  frequency perception module, high-resolution preserving, and correction fusion module, where `baseline' is the PVT backbone for camouflaged object detection. 
The comparison results are shown in Table \ref{tab:ablation}.

\begin{table}[t]
\caption{Quantitative results of ablation studies on the COD10k-Test dataset. First-stage and Second-stage mean the Frequency-guided Coarse Positioning and Detail-preserving Fine Localization respectively. HRP denotes High-res Preserving.}
\footnotesize{
\begin{tabular}{c|c | c | c | c llll}
\hline
\multirow{2}{*}{Baseline} &
\multicolumn{1}{c|}{1st-stage} & \multicolumn{2}{c|}{2nd-stage}  & \multicolumn{4}{c}{COD10k-Test (2026 images)}     \\ \cline{2-8}  
  & FPM & HRP & CFM & \multicolumn{1}{c}{\(S_\alpha\uparrow\) } & \multicolumn{1}{c}{ \(E_{mean} \uparrow\)} & \multicolumn{1}{c}{\(F_\beta^\omega \uparrow\)  } & \(M\downarrow\)       \\ \hline
    $\checkmark$ &  & & & 0.835                 & 0.899                  & 0.701                 & 0.032 \\
        $\checkmark$ & $\checkmark$ & & & 0.844                  & 0.908                 & 0.728                  & 0.031 \\
        $\checkmark$ & $\checkmark$ &$\checkmark$ & & 0.849                  & 0.911                  & 0.739                  & 0.030 \\
       $\checkmark$ & $\checkmark$ &$\checkmark$ & $\checkmark$ &\textbf{0.850 }                 & \textbf{0.913 }                 & \textbf{0.748}                  & \textbf{0.029} \\
\hline
\end{tabular}
}
\label{tab:ablation}
\end{table}

\noindent \textbf{Effectiveness of Frequency Perception Module.}
The proposed frequency perception module incorporates Octave convolution \cite{r40} that minimizes spatial redundancy and automatically learns high-frequency and low-frequency features. As shown in Table \ref{tab:ablation}, if we add the frequency perception module (\ie, baseline+FPM), all metrics can obtain performance gains compared with the PVT-alone without the octave convolution. The good performance lies that FPM learns 
rich frequency-aware information especially the high-frequency clues that are useful for camouflaged object coarse positioning. 
The other advantage of the FPM lies in that it is automatically online learning without any other extra offline operations. Thus, the flexibility and the high performance of the FPM make it suitable for accurately detecting camouflaged objects in real-world scenes. And FPM can also be easily integrated into other frameworks to assist in distinguishing the obscure boundaries of objects that are similar to the background.

\noindent \textbf{Effectiveness of High-resolution Preserving Module.}
Although The first frequency-guided coarse positioning stage (\ie, PVT+FPM) has achieved good target prediction maps, the object boundaries are still unsatisfactory. Thus, we adopt the high-resolution preserving mechanism for further detail refining. As shown in Table \ref{tab:ablation}, we conduct the detail-preserving fine localization stage without the correction fusion module (CFM) upon the first coarse positioning stage, \ie, PVT+FPM+High-res Preserving. If we introduce the low-level RGB-aware feature with high resolution to guide the refining process, we can find that the network outperforms the one PVT+FPM. The reason why we need a high-resolution preserving module for fine localization lies in two aspects, \ie, 1) the scales of camouflaged objects are various, and 2) the boundaries of camouflaged objects are usually meticulous which are hard to discern through the high-level semantic features. Inspired by the human visual perception system, humans usually need to zoom in on subtle details in a clear, high-resolution view to recognize camouflaged objects. If the scale is small in the image, we need to leverage the low-level edge-aware or shape-aware information to help the network obtain a fine localization. For the obscure boundary problem, multi-scale features fused in a step-by-step manner will give more help to the boundary separating from the complex background. Thus, we design the refining mechanism to integrate the high-resolution information and gradually fuse deep features together to solve these problems. 
The experimental results also show that the high-resolution preserving part can provide more performance gains for detail refining. 
And we can conclude that the detail refinement strategy is not only significant but also effective in localizing the camouflaged objects.

\noindent \textbf{Effectiveness of Correction Fusion Module.}
Though the high-resolution preserving mechanism for detail refining achieves good performance, the coarse camouflaged mask from the frequency-guided stage is still not effectively exploited enough. Thus, we propose a correction fusion module to further improve the quality of the camouflaged mask by completely mining the ability of the coarse map and the neighbor features. Specifically, we implement the CFM on the detail-preserving fine localization stage, the results are shown in the last row of Table \ref{tab:ablation}. While we update the detail-preserving with the CFM mechanism, all metric scores can be further improved especially in terms of the $S_\alpha$ and $F_\beta^\omega$ scores. The good target detection ability indicates that CFM plays an essential role in improving the detection performance of camouflaged objects. The main reason lies that CFM takes the prior coarse camouflaged mask and the neighbor layer interaction into account. First, the prior coarse prediction mask can provide us with an accurate region of the highlighted camouflaged objects which can extract object-centric related features well. 
Then, the channel-wise correlation correlates and combines neighbor layers to enhance the object representation which is more distinguishable for perceiving the camouflaged objects. 
Since CFM learns the channel correlation between adjacent features to obtain learnable weight maps and adjust original features, the dynamic mechanism achieves superior performance compared to the simple concatenation method (the third-row result of Table \ref{tab:ablation}). 
The good performance reflects that progressively fusing the prior coarse mask and cross-layer interaction is beneficial for camouflaged object refining.

\begin{figure}[!t]
\centering
\includegraphics[width=0.98\columnwidth]{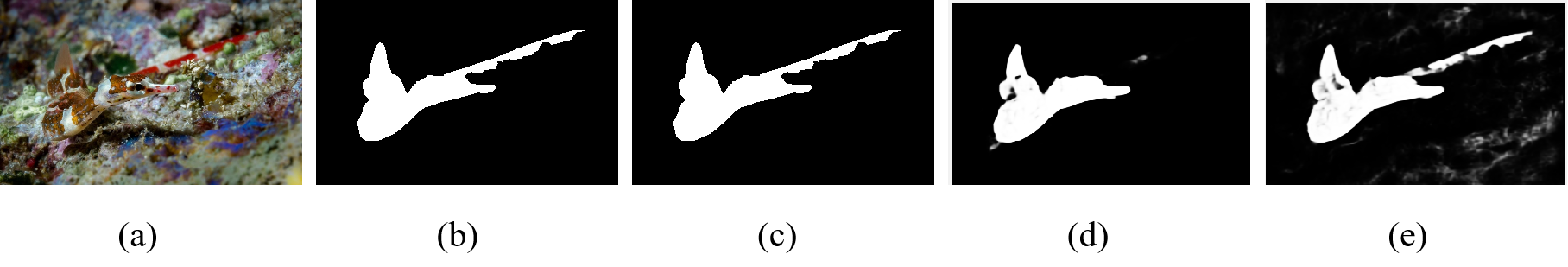}
\caption{Comparison results of different feature sets output from frequency perception module. (a) Input image. (b) GT. (c) Prediction result with our frequency fusing mechanism. (d) Prediction with only high-frequency features. (e) Result with only low-frequency features.}
\label{fig:fre1}
\end{figure}

In summary, the frequency-guided coarse positioning stage mainly highlights the important regions of the camouflaged objects under the guidance of hierarchy frequency-aware semantic information, and the detail-preserving fine localization stage further assists in separating the camouflaged objects from the obscure boundaries of the complex background by integrating the high-resolution clue, adjacent correlation features, and the coarse prediction mask. Finally, the proposed FPNet leads us to an accurate and effective solution for detecting camouflaged objects.

\subsubsection{\textbf{Detail Analysis of the Frequency-aware Information.}}
\begin{figure}[!t]
\centering
\includegraphics[width=1\columnwidth]{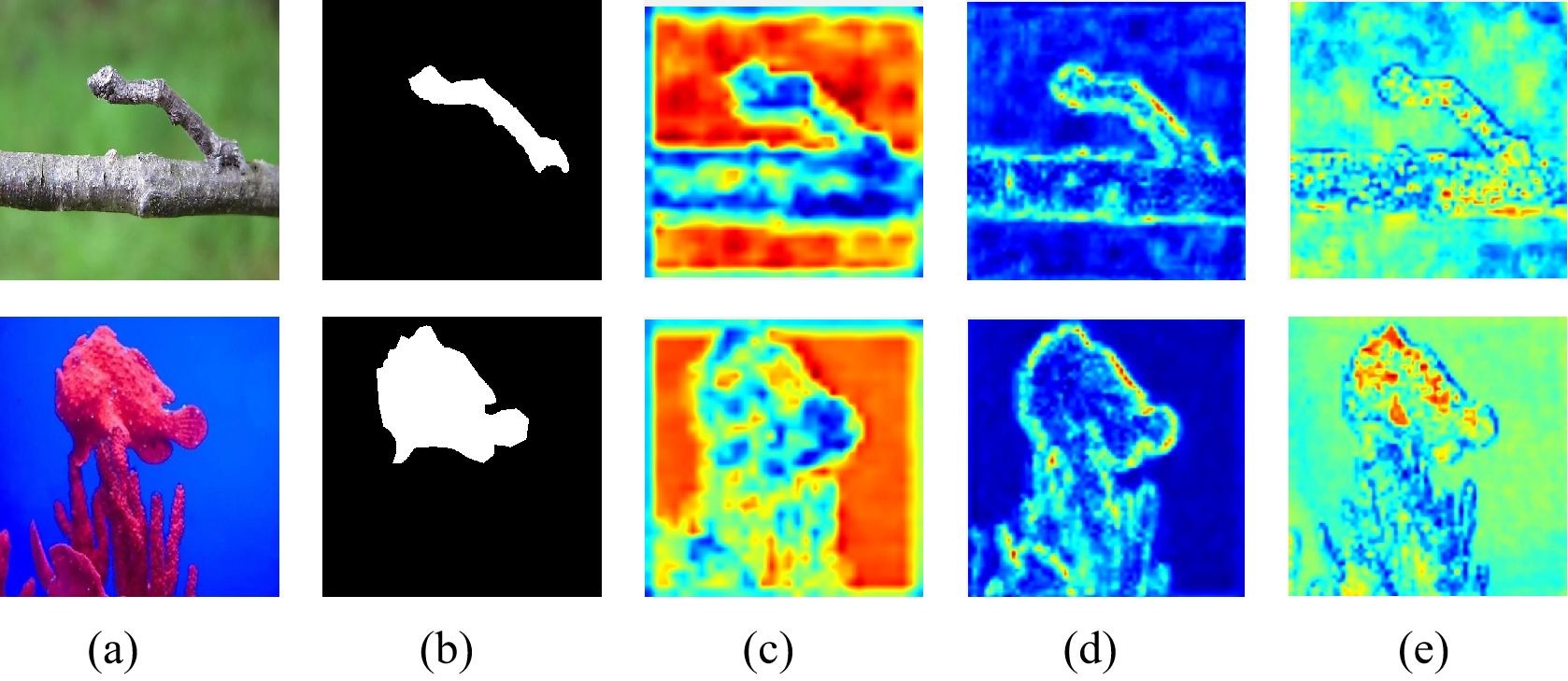}
\caption{Visualization of the learned features about the high-frequency and low-frequency groups. (a) Input image. (b) GT. (c) Low-frequency features. (d) High-frequency features. (e) Features fusing high-frequency and low-frequency after octave convolution.}
\label{fig:fre2}
\end{figure}

In order to verify the effectiveness of the frequency perception mechanism, we analyze different frequency fusing types through quantitative results, as shown in Table \ref{frequency}.
We also provide some visualization comparison results from the prediction mask and the learned frequency features, as shown in Figures \ref{fig:fre1} and \ref{fig:fre2}.

The proposed frequency perception module can automatically separate the features into high-frequency and low-frequency related features. However, how to choose a suitable way to integrate the prominent frequency-aware features to help obtain satisfactory camouflaged object masks needs further discussion. To verify it, we design different comparisons, \ie, only using high-frequency or low-frequency branches for following camouflaged object mask prediction. The detailed comparisons on the COD10k-Test dataset are shown in Table \ref{frequency}. We can observe that the only high-frequency method gives more help than the low-frequency method. All metrics scores of only high-frequency outperform the low-frequency to a great extent. This reflects that high-frequency information is more robust and distinguishable 
for recognizing camouflaged objects. It also meets the human visual system, we usually employ high-frequency clues to discern the target object from the uncertain region.  
However, since the octave convolution is an unsupervised operation, that is, no frequency-labeled maps by humans are used for optimization. Thus, some features learned from the low-frequency branch may be useful for camouflaged object detection. Moreover, we adopt a simple addition operation fusing the high-frequency and low-frequency together, the result is shown in the last row of Table \ref{frequency}. The simple addition of high and low frequencies achieves the best performance over only single-frequency ones. Based on these observations, we suggest combining the high-frequency and low-frequency into a single addition to obtain further improvements. 
\begin{table}[!t]
\caption{Quantitative results of frequency-aware features on the COD10k-Test dataset.}
\begin{tabular}{cc|cccc}
\hline
\multirow{2}{*}{Ver.} & \multirow{2}{*}{Method} & \multicolumn{4}{c}{COD10k-Test (2026 images)}                                                          \\ \cline{3-6} 
                      &                         & \multicolumn{1}{l}{\(S_\alpha\uparrow\) } & \multicolumn{1}{l}{\(E_{mean} \uparrow\)} & \multicolumn{1}{l}{\(F_\beta^\omega \uparrow\)} & \multicolumn{1}{l}{\(M\downarrow\) } \\ \hline
No.1                  & Low-frequency           & 0.765                 & 0.831                 & 0.438                 & 0.061                 \\
No.2                  & High-frequency          & 0.848                 & 0.910                 & 0.747                 & 0.029                 \\
No.3                  & High-fre+Low-fre          & 0.850                 & 0.913                 & 0.748                 & 0.029                 \\ \hline
\end{tabular}
\label{frequency}
\end{table}

In Figure \ref{fig:fre1}, we analyze the influence of different frequency-aware feature types. In particular, the prediction masks of only high-frequency, only low-frequency, and ours (high-frequency+low-frequency) are shown. The high-frequency method can predict the key part of the camouflaged object, the low-frequency method can obtain an intact region but with some interference background regions. 
Our proposed method can obtain an accurate object mask compared with the high-frequency or low-frequency ones. The comparison results indicate that the frequency features are meaningful for camouflaged object detection. And fusing the high-frequency and low-frequency will further assist the model in obtaining a relatively complete object mask. 

We also visualize the learned frequency-aware features via the octave convolution to further explain the effectiveness of the proposed frequency perception mechanism, as shown in Figure \ref{fig:fre2}. 
First, our proposed frequency perception mechanism can automatically separate the frequency features into high and low frequency groups without any frequency supervision information.
Second, we can observe that the high-frequency and low-frequency groups in the learning process of octave convolution extract the edge information and the main part of the image, respectively. 
The low-frequency group (Figure \ref{fig:fre2}(c)) focuses more on the overall composition of the image, while the high-frequency group (Figure \ref{fig:fre2}(d)) portrays the edge part of the camouflaged object in the image.
While combing the low-frequency and high-frequency groups (Figure \ref{fig:fre2}(e)), our model can focus on the crucial regions of the camouflaged object despite it is similar to the surrounding region. 

In conclusion, the proposed frequency perception network has been verified by analyzing the qualitative and quantitative comparison results that the frequency information can give more help to camouflaged object detection. And the proposed frequency perception module can be plugged and played into arbitrary frameworks.

\section{Conclusion}
In this paper, we propose a frequency-perception network (FPNet) to address the challenge of camouflaged object detection by incorporating both RGB and frequency domains. Specifically, a frequency-perception module is proposed to automatically separate frequency information leading the model to a good coarse mask at the first stage. Then, a detail-preserving fine localization module equipped with a correction fusion module is explored to refine the coarse prediction map. Comprehensive comparisons and ablation studies on three benchmark COD datasets have validated the effectiveness of the proposed FPNet. 
This work will benefit more sophisticated algorithms exploiting frequency clues pursuing appropriate solutions in various areas of the multimedia community. In addition, the long-tail problem also exists in COD, this motivates us to explore reasonable solutions referring to the typical methods of long-tail recognition \cite{yang2021learning, yang2022optimizing}.

\begin{acks}
This work was supported in part by the National Key R\&D Program of China under Grant 2021ZD0112100, in part by the National Natural Science Foundation of China under Grant 62002014, Grant 62202461, Grant 62271180, Grant U1913204, Grant U1936212, Grant 62120106009, in part by the Taishan Scholar Project of Shandong Province under Grant tsqn202306079, in part by Young Elite Scientist Sponsorship Program by the China Association for Science and Technology under Grant 2020QNRC001, in part by the Project for Self-Developed Innovation Team of Jinan City under Grant 2021GXRC038, in part by CAAI-Huawei MindSpore Open Fund, and in part by China Postdoctoral Science Foundation under Grant 2022M723364.

\end{acks}

\clearpage

\bibliographystyle{ACM-Reference-Format}
\bibliography{acmart}

\newpage
\appendix
\section{Appendix}
\subsection{Qualitative Comparison with SOTA Methods}
\begin{figure*}[!t]
\centering
\includegraphics[width=0.6\linewidth]{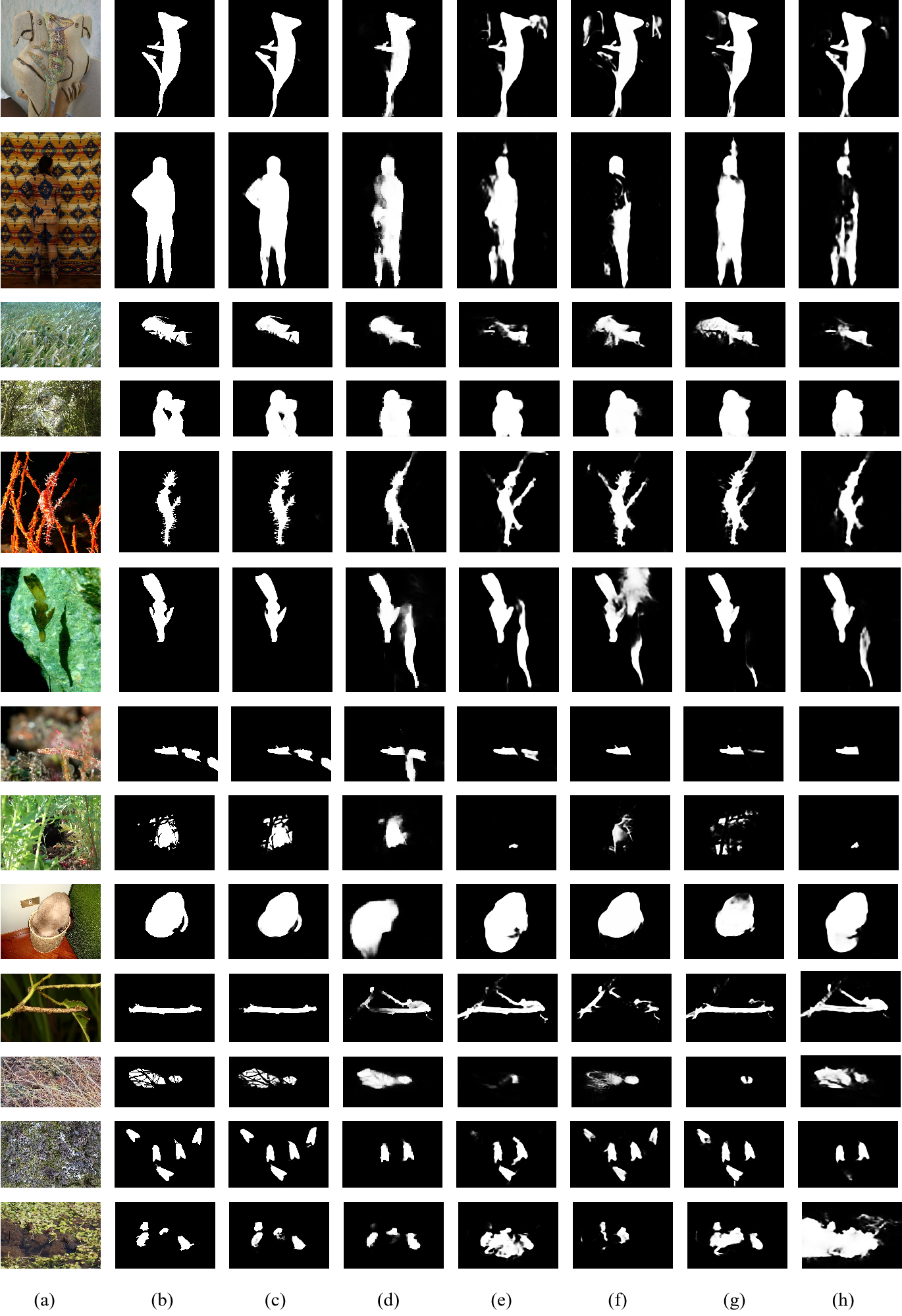}
\caption{Qualitative results of our proposed FPNet model and some state-of-the-art COD methods. The images from left to right are (a) Input image, (b) GT, (c) Ours, (d) FreNet \cite{r39}, (e) SINet-V2 \cite{r53}, (f) PFNet \cite{r15}, (g) LSR \cite{r38}, and (h) PraNet \cite{r24}.}
\label{fig:app1}
\end{figure*}

In Figure \ref{fig:app1}, we provide more visual examples of different methods. We can see that our proposed network is still competitive in challenging and difficult scenarios, such as multiple objects, fine objects and complex background distractions.

When there are multiple camouflaged objects, as shown in the last two rows of Figure \ref{fig:app1}, other SOTA methods either cannot identify all the camouflaged objects well or the boundaary of the recognized object are not clear. 
In contrast, our proposed FPNet network can predict all the camouflaged objects with clear and sharp boundary. 

The detection results of our model have a clear advantage in describing the fine details of camouflaged objects. For example, the camouflaged object  in the fifth image of Figure \ref{fig:app1} has many small, thin, burr-like structures. Compared with other methods, only our method not only accurately detects the camouflaged object, but also fully characterizes these trivial details. 
Similar advantages are also reflected in the situation that the camouflaged target is occluded.
For example, in the images seventh, eighth, eleventh rows of Figure \ref{fig:app1}, the camouflaged object is partially occluded, but our method is still robust to this case. It is worth mentioning that we also accurately exclude non-camouflaged occluded object regions from the prediction results.


Furthermore, our method is also able to handle challenging complex background scenes, such as the first, second, sixth, ninth and tenth rows of Figure \ref{fig:app1}. 
Taking the sixth row image as an example, we should detect a ghost pipefish from the image. Other methods treat the indistinguishable shadows of the ghost pipefish as camouflaged object, while only our method can effectively detect the ghost pipefish and eliminate interference from shadows.

\subsection {Visualization of Ablation Studies}
\begin{figure*}[!t]
\centering
\includegraphics[width=0.55\linewidth]{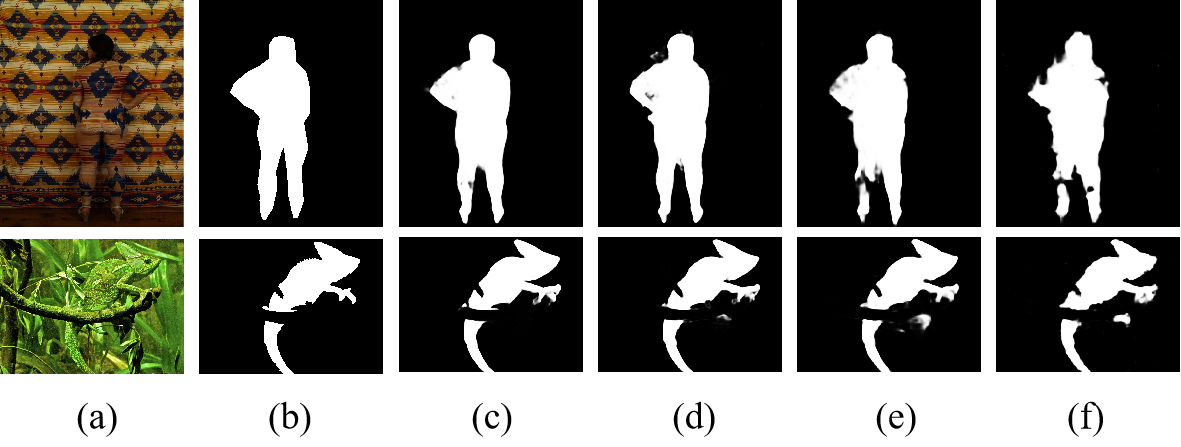}
\caption{Qualitative results of ablation studies. The images from left to right are (a) Input image, (b) GT, (c) Ours, (d) High-res Preserving, (e) Frequency-guided Coarse Positioning, (f) Baseline. }
\label{fig:app2}
\end{figure*}

We also supplement the visualization results of the ablation experiment in Figure \ref{fig:app2}. 
Taking the first image as an example, our baseline model (\ie, Figure \ref{fig:app2}(f)) can only roughly determine the main part of the camouflaged object, and there is still much room for improvement in terms of details and accuracy.
Further, after the introduction of the frequency-perception module in the baseline model, the left shoulder area of the camouflaged human has been significantly improved, but the problem of leg integrity remains unresolved.
Then, we add a high-resolution preserving design to our network, which makes the leg details more complete but introduces some noise.
Finally, using our designed correction fusion module in our network allows us to achieve the best performance with accurate location, complete structure, and sharp boundary.


\end{document}